\documentclass{article}
\pdfoutput=1

\usepackage{todonotes}
\usepackage{subcaption}
\usepackage{multirow}
\usepackage[toc,page]{appendix}

\usepackage[final,nonatbib]{nips_2018}

\usepackage[utf8]{inputenc} 
\usepackage[T1]{fontenc}    
\usepackage{hyperref}       
\usepackage{url}            
\usepackage{booktabs}       
\usepackage{amsfonts}       
\usepackage{nicefrac}       
\usepackage{microtype}      
\usepackage{graphicx}
\usepackage[sort,numbers]{natbib}

\title{Migration through Machine Learning Lens - Predicting Sexual and Reproductive Health Vulnerability of Young Migrants}

\author{%
    Amber Nigam\\
  Georgia Institute of Technology\\
  Atlanta, GA 30332, United States \\
  \texttt{anigam9@gatech.edu} \\
  \And
  Pragati Jaiswal\\
  Harvard T.H. Chan School of Public Health\\
  Boston, MA 02115, United States\\
  \texttt{pragatijaiswal3190@gmail.com} \\
  \AND
    Teertha Arora\\
  Harvard T.H. Chan School of Public Health\\
  Boston, MA 02115, United States\\
  \texttt{tarora@hsph.harvard.edu} \\
  \And
  Uma Girkar\\
  Massachusetts Institute of Technology\\
  Cambridge, MA 02139, United States\\
  \texttt{umag@mit.edu, umag@google.com} \\
  \And
  Leo A. Celi\\
  Laboratory for Computational Physiology, Massachusetts Institute of Technology\\
  Cambridge, MA 02139, United States\\
  \texttt{lceli@mit.edu} \\
}

\begin{document}

\maketitle

\begin{abstract}
In this paper, we have discussed initial findings and results of our experiment to predict sexual and reproductive health vulnerabilities of migrants in a data-constrained environment. Notwithstanding the limited research and data about migrants and migration cities, we propose a solution that simultaneously focuses on data gathering from migrants, augmenting awareness of the migrants to reduce mishaps, and setting up a mechanism to present insights to the key stakeholders in migration to act upon. We have designed a webapp for the stakeholders involved in migration: migrants, who would participate in data gathering process and can also use the app for getting to know safety and awareness tips based on analysis of the data received; public health workers, who would have an access to the database of migrants on the app; policy makers, who would have a greater understanding of the ground reality, and of the patterns of migration through machine-learned analysis. Finally, we have experimented with different machine learning models on an artificially curated dataset. We have shown, through experiments, how machine learning can assist in predicting the migrants at risk and can also help in identifying the critical factors that make migration dangerous for migrants. The results for identifying vulnerable migrants through machine learning algorithms are statistically significant at an alpha of 0.05.
\end{abstract}

\section{Introduction and Related Work}
One of the major reasons why people have increasingly moved across borders in the last few decades is due to factors like socioeconomic conditions in their country of origin and climate change, in search of a better future for themselves as well as their families. During such transit, many witness exploitation or trauma which may affect their physical and mental well-being. Further, access to healthcare in a new environment for migrants may be restricted when compared to the local residents. This can be due to either inadequate coverage of healthcare services for migrant population or lack of awareness regarding clinic locations. In the scenario of migration, the ‘adolescent’ population is particularly vulnerable and fragile. Not having complete autonomy over their decisions due to financial and social dependence, they are not empowered enough to take decisions for their Sexual and Reproductive Health (SRH). As a result, their SRH gets compromised. The barriers to health services get intensified due to inadequate sources of information, lack of financial resources and paucity of youth-friendly health services.

A study by \cite{bocquier2011migrant} that examined the impact of mother and child migration on the survival of more than 10,000 children in two of Nairobi’s informal settlements between 2003 and 2007 found that children born to women who were pregnant at the time of migration have the highest risk of dying. Another study by \cite{greif2011internal} explored the vulnerability of the migrant population to engage in risky sexual encounters and it was found that migrant populations are more prone to engage in risky sexual behavior.

\begin{figure}
    \centering
    \includegraphics[width=290pt]{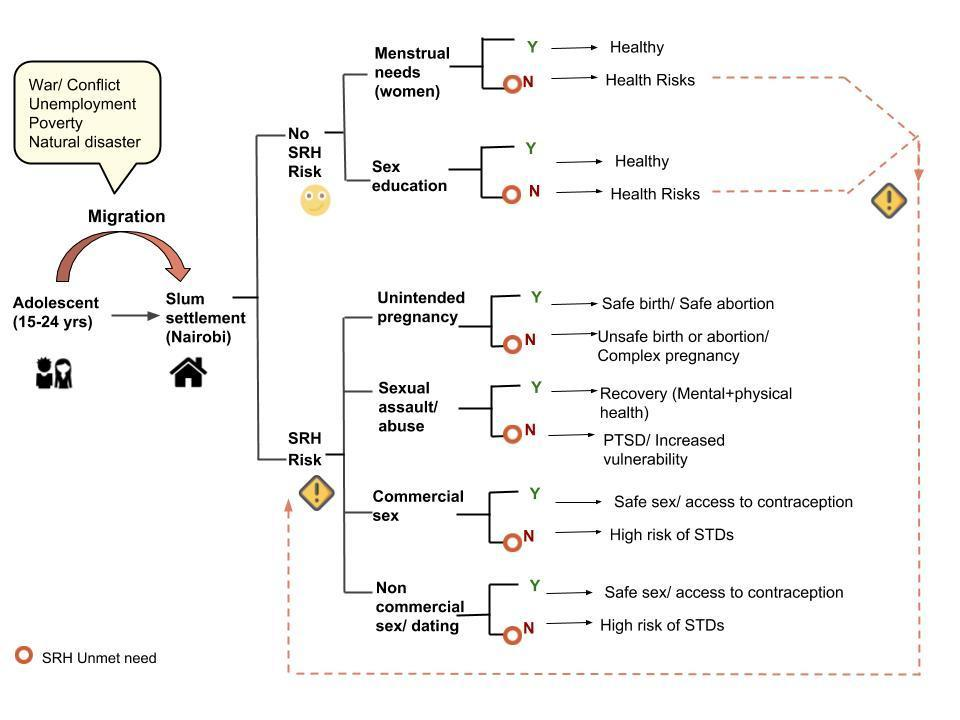}
    \caption{Mapping potential areas for SRH needs for an adolescent migrant.}
    \label{fig:my_label}
\end{figure}

These concerns prompted UNFPA-MIT team to explore the needs of migrant youth as well as identifying the barriers to existing health services in the migrated country. To understand the key element of our user base – adolescent migrants – we mapped their geographical journey (Figure \ref{fig:my_label}) to identify potential areas where SRH services will be needed. For instance, if an adolescent migrant is exposed to harassment then access to SRH services might help them recover swiftly both mentally and physically. On the other side, in case of unintended pregnancy, there are increased chances of unsafe birth or abortion if requisite reproductive health services are inaccessible.

Finally, we experimented with machine learning on artificially created data to come up with recommendations on vulnerability of migrants for maximizing attack surface against the problems faced by migrants. This could not only help in prioritizing preventive actions, but it could also help in formulating a data-driven policy and also lead the way for dealing with such issues at scale.

\section{Methods}
 
When young people migrate, many a time they lack awareness of their rights, and the fact they are in a foreign territory further adds to the burden on their sexual and reproductive health. The intent of our experiment is to figure out ways of making migration safe through: 1. providing tips that might help them deal with situations independently or seek help whenever required 2. helping the migrants to be aware of their rights 3. identifying vulnerable migrants 4. identifying factors causing vulnerability.

We created a webapp for providing migration-based information, like safety tips, to the migrants and for collecting information from the migrants that could be used for deriving insights later using machine learning. The website would also be useful for other stakeholders like researchers, healthcare workers and policy makers as it would provide information collected from the migrants for their respective use-cases.

We also developed a recommender system for predicting vulnerable migrants for identifying the most predictive features, both migrant-based and geography-based, leading to the vulnerability. The vulnerability of a migrant is the increased possibility of risk to their sexual and reproductive health because of factors like lack of awareness and hostile living conditions of the migrating city, among other factors. The recommender system would help tag vulnerable migrants and would serve as a prioritizing tool for healthcare workers and policy makers in addressing challenges that arise due to migration.

\subsection{Dataset - Preparing Pipeline of Data Collection from Migrants and Manual Curation of Migration Dataset for Machine Learning Experiments}

We shortlisted 27 questions to be administered by the webapp on topics relating to profile screening, migration background, sexual and reproductive health knowledge levels as well as medical history to ascertain the needs of the migrant youth population in Central Asia and Africa. A subset of these questions would help find the required features (see Table \ref{Table 1}) for migrants and migration cities for the machine learning analysis. Next, given we were working on the problem from scratch, we did not have any readily available dataset on which to base our experiments, we decided to manually curate a rule-based dataset to address the cold-start. The rules used for training, validation and testing sets were defined independently by different members of the group to mimic a real-world like scenario. Although manually-curated data is not comparable to real-world data, it helped us frame a quick proof-of-concept to predict vulnerability of the migrants using machine learning.

\begin{table}
  \caption{Features and Their Value Types}
  \label{Table 1}
  \centering
  \begin{tabular}{ll}
    \toprule
    Feature     & Value Type\\
    \midrule
    Age & Integral     \\
    Sex     & Categorical      \\
    City of Birth     & Categorical  \\
    Current / Migration City     & Categorical  \\
    Duration of stay in current City (in months)     & Integer  \\
    Married, divorced, widowed     & Categorical  \\

    \bottomrule
  \end{tabular}
\end{table}

\subsection{Models - Machine Learning Analysis for Classifying Vulnerable Migrants}

We used different machine learning algorithms like Random Forest \cite{breiman1996bagging, breiman2001random, liaw2002classification}, Support Vector Machine (SVM) \cite{cortes1995support}, XGBoost \cite{chen2016xgboost} and Sequential Neural Network \cite{hagan1996neural} to predict vulnerability (whether one suffered any physical abuse) based on the features of migrants and migrating cities. We believe that the most important aspect of solution for the problem we are trying to solve is to not miss predicting vulnerable migrants (i.e. high recall for vulnerable class is needed) even if it results in a slight loss of precision for vulnerable class (i.e can afford a few false-positives). Therefore, we have optimized our algorithms to minimize false-negatives (although, we could have settled for a much higher F1 score). As we can see in Table \ref{Table 2}, we have correctly identified the majority of the vulnerable migrants and also reduced the number of migrants to be checked on priority at a cost of a few false positives. For instance, we are able to correctly identify 14 out of 17 vulnerable migrants and reduce the number of people to monitor on priority from 200 (total migrant count) to 28 (positively predicted cases i.e. true positive + false positive) at a cost of 14 false positive instances using SVM algorithm.

We have also evaluated feature importance using Random Forest and found “age of migrant” and “accompanying adult in the family” to be the top predictive features for our dataset. We acknowledge that rules used to build the dataset for this experiment are curated manually and the patterns in real world scenario would be much more convoluted. But the intent of this exercise is to demonstrate, through a simple rule-based dataset, how machine learning could identify the patterns that could exist in manually-curated or real-world dataset. We used keras and sklearn libraries for our machine learning and deep learning implementation. The proportion between training, validation testing datasets is 70:15:15. The intent of this experiment is to show how machine learning can help extract underlying rules / patterns and help in determining vulnerabilities. This could help in prioritizing actions to save as many people as possible.

\section{Results}
In this paper, we have described different ways of disseminating migration-based information to the migrants through webapp. The webapp also serves as a utility for gathering information about migrants and for providing analytics on the information received to the healthcare workers and policy makers. We have also created a proof-of-concept for predicting vulnerability of the migrants that would help in identifying vulnerable migrants and the critical factors in migration that can help it make safer, like places to avoid for migrants and the importance of an accompanying adult in the family. In our curated dataset, “age of the migrant” and “accompanying adult in the family” were the two most predictive features in identifying vulnerability of the migrant, considering feature importance computed using Random Forest. The F1 scores, accuracy scores and confusion matrices for different algorithms evaluated are shown in Table \ref{Table 2}. Our results also show that, for our manually-curated dataset, we have a high recall for vulnerable migrants class (i.e. correctly identified most of the vulnerable migrants) and have also reduced the number of migrants to be checked on priority.

\begin{table}
  \caption{F1 Score and Accuracy}
  \label{Table 2}
  \centering
  \begin{tabular}{lllllll}
    \toprule
Algorithm & F1 Score & Accuracy & TN & FP & FN & TP \\
\midrule
SVM     & 0.62     & 0.92     & 169     & 14     & 3     & 14 \\
Random Forest     & 0.61     & 0.90     & 166     & 17     & 2     & 15 \\
XGBoost     & 0.59     & 0.89     & 162     & 21     & 1     & 16 \\
MLP     & 0.60     & 0.92     & 170     & 13     & 4     & 13 \\
Sequential Neural Network     & 0.50     & 0.89     & 168     & 5     & 15     & 10 \\
    \bottomrule
  \end{tabular}
\end{table}

\section{Conclusion and Future Work}
For addressing problems such as the current one, where machine learning can only be used when there is data, we have shown that such a cold-start can be addressed by artificially curating the data using rules. Later, when the data is collated, it can be used to train algorithms for providing a more comprehensive and nuanced solution.

In the future, we will conduct an exhaustive geospatial analysis to locate the migrant population. We primarily looked at urban areas for the scope of this project as we expected a greater concentration of migrants in those regions; however subsequent research could focus on data collection for at-risk migrants living in rural communities or dispersed in small numbers across various regions. The start and end locations of the migrants can also be recorded to look for migration patterns among specific groups of people and to investigate the extent to which starting locations influence final destinations.

An integral next step will be to publicize the utility of the app among healthcare providers. If both migrants and healthcare providers are actively using the app, a telemedicine-based approach could be used to diagnose and treat patients remotely without them having to even come into the healthcare centers except for urgent issues. This would save both the migrants and the healthcare providers much time and money and could be especially beneficial for migrants living in remote locations or far away from healthcare centers. In the long term, we seek to have a strong patient-provider network based on this app for migrants. This integration would sensitize the healthcare providers towards the migrants’ needs and possibly create treatment approaches most specific to them.

Finally, we have shown that machine learning and deep learning algorithms are able to identify most of the vulnerable migrants, albeit in an artificially curated dataset, at a cost of a few false positives. We acknowledge that rules used to build the dataset for this experiment are curated manually and the patterns in real world scenario would be much more convoluted. But the intent of this exercise is to demonstrate, through a simple rule-based dataset, how machine learning could identify the patterns that could exist in manually-curated or real-world dataset. The next logical step would be to run these algorithms over the actual data. It would also be interesting to predict the severity and probability of abuse through the algorithms.
\bibliographystyle{unsrt}
\bibliography{main}

\end{document}